\documentclass{article} 
\usepackage{iclr2024_conference,times}

\usepackage{amsmath,amsfonts,bm}

\def\eqref#1{equation~\ref{#1}}

\def\1{\bm{1}}

\DeclareMathAlphabet{\mathsfit}{\encodingdefault}{\sfdefault}{m}{sl}
\SetMathAlphabet{\mathsfit}{bold}{\encodingdefault}{\sfdefault}{bx}{n}

\usepackage{hyperref}
\usepackage{url}
\usepackage{graphicx}
\usepackage{adjustbox}
\usepackage[utf8]{inputenc} 
\usepackage[T1]{fontenc}    
\usepackage{hyperref}       
\usepackage{url}            
\usepackage{booktabs}       
\usepackage{amsfonts}       
\usepackage{nicefrac}       
\usepackage{microtype}      
\usepackage{xcolor}         
\usepackage{lipsum}
\usepackage{graphicx}
\usepackage{color}
\usepackage{amsmath}

\title{Diffusion-Promoted HDR Video Reconstruction}

\author{Yuanshen Guan\textsuperscript{1} \quad Ruikang Xu\textsuperscript{1} \quad Mingde Yao\textsuperscript{2} \quad Ruisheng Gao\textsuperscript{1}
\\ \textbf{Lizhi Wang\textsuperscript{3} \quad Zhiwei Xiong\textsuperscript{1}} \\
\textsuperscript{1}University of Science and Technology of China \\
\textsuperscript{2}The Chinese University of Hong Kong \quad 
\textsuperscript{3}Beijing Institute of Technology 
}

\iclrfinalcopy 
\begin{document}

\maketitle

\begin{abstract}

High dynamic range (HDR) video reconstruction aims to generate HDR videos from low dynamic range (LDR) frames captured with alternating exposures. Most existing works solely rely on the regression-based paradigm, leading to adverse effects such as ghosting artifacts and missing details in saturated regions. In this paper, we propose a diffusion-promoted method for HDR video reconstruction, termed HDR-V-Diff, which incorporates a diffusion model to capture the HDR distribution. As such, HDR-V-Diff can reconstruct HDR videos with realistic details while alleviating ghosting artifacts. However, the direct introduction of video diffusion models would impose massive computational burden. Instead, to alleviate this burden, we first propose an HDR Latent Diffusion Model (HDR-LDM) to learn the distribution prior of single HDR frames. Specifically, HDR-LDM incorporates a tonemapping strategy to compress HDR frames into the latent space and a novel exposure embedding to aggregate the exposure information into the diffusion process. We then propose a Temporal-Consistent Alignment Module (TCAM) to learn the temporal information as a complement for HDR-LDM, which conducts coarse-to-fine feature alignment at different scales among video frames. Finally, we design a Zero-Init Cross-Attention (ZiCA) mechanism to effectively integrate the learned distribution prior and temporal information for generating HDR frames. Extensive experiments validate that HDR-V-Diff achieves state-of-the-art results on several representative datasets.

\end{abstract}
\section{Introduction}

High dynamic range (HDR) content offers viewers a visually rich experience with its ability to showcase high contrast and a wide array of colors.
As video-on-demand services make diverse videos readily accessible, the demand for HDR video content is on the rise.  Despite the widespread availability of HDR displays, there remains a notable scarcity of HDR content available for deployment.

HDR video reconstruction provides a practical way to generate HDR video from low dynamic range (LDR) frames captured with alternating exposures~\citep{kala13,kala2014hdr,iccv21hdr,lanhdr}, which builds upon and extends the principles of multi-exposure fusion~\citep{hdrimage04,hdrimage05,hdrimage06, AHDRNet,hdrt}. 
Specifically, the missing visual details in under-exposed and over-exposed regions at a specific time step can be effectively found in the adjacent frames with other exposures.
Conventional HDR video reconstruction methods~\citep{mangiat2010high,mangiat2011spatially,kala13,kala2014hdr} conduct motion compensation with the estimated motion prior (\textit{e.g.}, optical flow~\citep{flow01,flow02}) before the merging process, inevitably producing artifacts due to inaccurate alignment with long optimization times.
Recent methods follow a regression-based learning paradigm~\citep{kala2014hdr,iccv21hdr,lanhdr}. 
Kalantari~\emph{et al.}~\citep{kala2014hdr} conduct flow alignment before a merging network.
%
Chen~\emph{et al.}~\citep{iccv21hdr} performs a more detailed two-stage alignment with explicit optical flow and deformable convolutions.
Chung~\emph{et al.}~\citep{lanhdr} propose a luminance-based alignment to register motions between frames by an implicit attention score.

\begin{figure*}[t]
  \centering
  \includegraphics[width=0.98\linewidth]{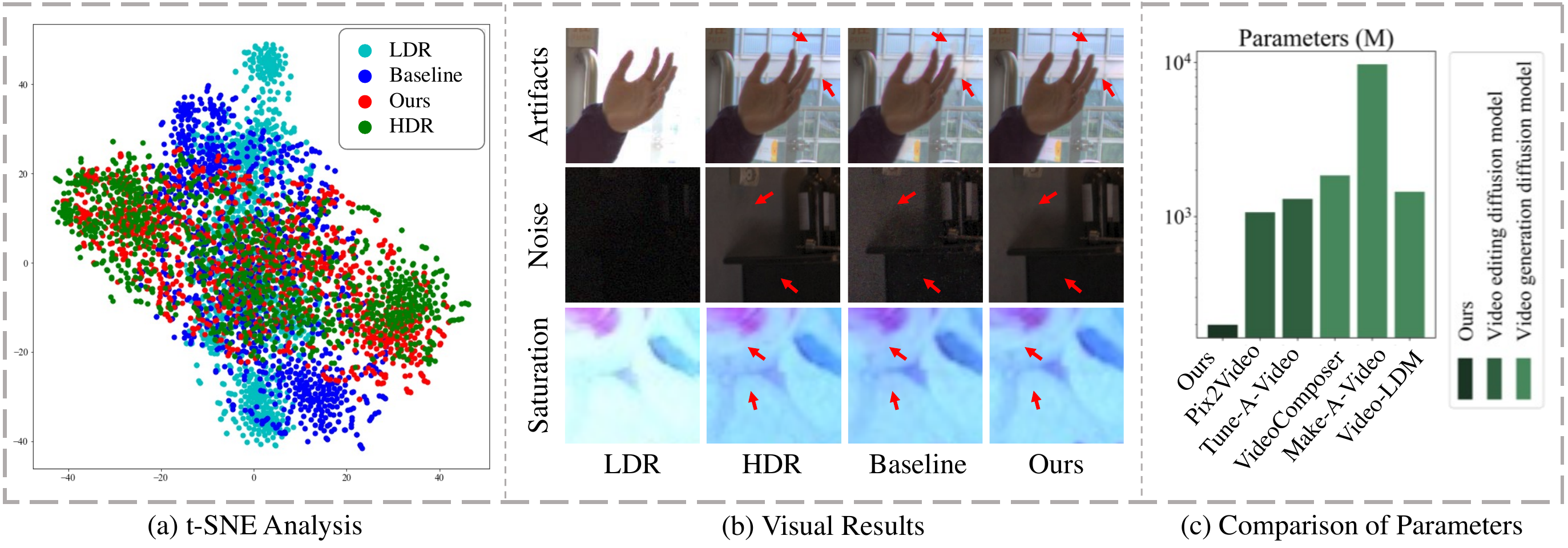}
  \vspace{-0.41cm}
   \caption{Motivation and visual results. 
   (a) illustrates the t-SNE distribution of input LDR frames, HDR frames, results of the regression-based baseline~\citep{lanhdr}, and our results. Each point represents a 256$\times$256 cropped patch.
   The output of the baseline falls outside the distribution of HDR frames, whereas our results are significantly closer to ground truth.
   (b) specifically highlights the differences in reconstruction quality.
   Results from the regression-based baseline~\citep{lanhdr} show artifacts and noise since its supervision strategy ignores the distribution learning.
   In contrast, our method exhibits fewer artifacts and noise with more reasonable details.
   (c) depicts parameter comparisons among the proposed method and video generation diffusion models, such as Make-A-Video~\citep{makevideo}, VideoComposer~\citep{videocomposer}, and Video-LDM~\citep{videoldm}. It also compares these with video editing diffusion models, including Tune-A-Video~\citep{tuneavideo} and Pix2Video~\citep{related_Ceylan}.}
   \label{fig:Teaser}
   \vspace{-0.72cm}
\end{figure*}

However, the optimization strategy of the regression-based paradigm  (\textit{i.e.}, $\mathcal{L}_1$ or $\mathcal{L}_2$ loss) only focuses on pixel-wise reconstruction quality, neglecting to capture the HDR distribution.
As depicted in Fig.~\ref{fig:Teaser} (a), the distribution of reconstructed results by a regression-based baseline~\citep{lanhdr} exhibits a large margin from the distribution of HDR video frames.
This problem leads to artifacts in fast-moving scenes, and unsatisfactory results in extremely dark and saturated regions, as shown in Fig.~\ref{fig:Teaser} (b).
This inherent limitation of the regression-based paradigm motivates us to learn the HDR distribution for achieving high-quality HDR video reconstruction.


In this paper, we propose a diffusion-promoted method for HDR video reconstruction, termed HDR-V-Diff, which incorporates a diffusion model to capture the HDR distribution.
Specifically, the diffusion model is a breakthrough paradigm in the area of generation~\citep{diff_image_01,diff_image_02,diff_image_03,diff_reso_01,diff_reso_02,diff_reso_03,diff_video_01,diff_video_02}, which aims to convert a simple known distribution (\textit{e.g.}, Gaussian) into a target distribution using a diffusion process.
By leveraging the distribution learning capability of the diffusion model, HDR-V-Diff can effectively approximate the distribution of HDR video frames, thereby reconstructing results with realistic details while alleviating ghosting artifacts, as shown in Fig.~\ref{fig:Teaser} (a) and (b).

Nevertheless, it is challenging to apply the diffusion model to HDR video reconstruction, since the existing video diffusion methods need large-scale models to achieve video generation as shown in Fig.~\ref{fig:Teaser}~(c).
However, employing large-scale models in this reconstruction task would result in a massive waste of computational resources, as most pixels and information in LDR videos are known and repeated in every frame.
To address this problem, we propose an HDR Latent Diffusion Model (HDR-LDM) to capture the distribution of single HDR frames under the condition of the corresponding LDR frames.
Specifically, HDR-LDM introduces a tonemapping strategy to compress HDR frames to the latent space, while utilizing a novel exposure embedding to aggregate the exposure value of LDR frames into the diffusion process. 
We then proposed a Temporal-Consistent Alignment Module (TCAM) to learn the temporal information as a complement for HDR-LDM, which performs coarse-to-fine feature alignment, \textit{i.e.}, spatial attention~\citep{AHDRNet,hdrt}, deformable convolution~\citep{edvr,iccv21hdr}, and patch matching~\citep{TTSR,lanhdr} at different scale. 
Finally, we introduce a novel Zero-init Cross-Attention (ZiCA) mechanism to integrate the distribution prior from HDR-LDM and the temporal information from TCAM.
Specifically, ZiCA utilizes the cross-attention mechanism to adaptive information fusion while introducing the zero-init convolution to stabilize optimization.


Our main contributions are as follows:
1) We propose a diffusion-promoted method for HDR video reconstruction that incorporates the generative paradigm to capture the distribution of HDR frames.
2) We propose HDR-LDM to produce the distribution prior of single HDR frames, and TCAM to learn the temporal information as a complement for HDR-LDM. We then design ZiCA to effectively integrate the learned distribution prior and temporal information.
3) Extensive experiments demonstrate that our method achieves state-of-the-art results on several benchmarks. To the best of our knowledge, our work is the first effort to apply the diffusion model for HDR video reconstruction.

\section{Related Work}
\paragraph{HDR Video Reconstruction}
HDR video reconstruction aims to restore HDR frames from alternating exposed LDR frames. 
Conventional methods~\citep{kang2003high,kala13,li2016maximum} typically involve motion compensation to mitigate artifacts before the merging process. 
These methods are always time-cost during the optimization process and yield unreliable alignment in severely misaligned scenes.
Recent methods~\citep{kalantari2019deep, iccv21hdr, lanhdr} utilize a regression-based paradigm, comprised of a flow-based or implicit feature alignment and a merging process. 
However, the optimization strategy of regression-based methods (\textit{i.e}. $\mathcal{L}_1$ or $\mathcal{L}_2$ loss) focuses on narrow pixel-wise reconstruction but neglects to capture the HDR distribution. This inherent shortage limits their quality and performance.
In this paper, we propose to leverage the diffusion model to capture the distribution of HDR frames, thereby promoting the quality of reconstructed videos.
\vspace{-2mm}
\paragraph{Video Diffusion Models}
Recently, Diffusion models have gradually unveiled their potential across various vision tasks~\citep{rombach2022high,diff_image_01,diff_image_02,diff_reso_01,diff_reso_02,diff_video_01,diff_video_02,videoldm,related_Ceylan,khachatryan2023text2video,zerowang2023,zeroinfusion}, encompassing the domain of video-related research, such as video generation~\citep{diff_video_01,diff_video_02,rombach2022high} and editing~\citep{esser2023structure,related_Ceylan}.
It can be observed that most current video-related diffusion models choose to utilize Latent Diffusion Model (LDM)~\citep{stablediffusion} to avoid heavy training and inference costs, and many of them~\citep{videoldm,related_Ceylan,khachatryan2023text2video,zeroinfusion,zerowang2023} choose to leverage priors from pretrained image diffusion models to achieve video generation or editing.
Despite the remarkable achievements of prior research, it is still challenging to employ diffusion models in video reconstruction tasks while avoiding unnecessary computational burden.
Moreover, HDR data distribution differs significantly from that of standard natural images.
As an exploratory  work, we propose to capture the distribution of single HDR frames with low computational cost by using LDM.
Additionally, we design an exposure embedding and a tonemapping strategy to enhance the effectiveness of the diffusion model towards HDR data.

\section{Proposed Method}
HDR video reconstruction aims to reconstruct high-quality HDR video frames from LDR videos with alternating exposures. 
Before feeding the LDR frames into the proposed HDR-V-Diff, we first map the LDR frames to the linear domain using gamma correction with $\gamma = 2.2$ following previous works~\citep{iccv21hdr,lanhdr}. Then $k$-th LDR frame $I_k$ and the corresponding transformed frame are concatenated along the channel dimension to construct a 6-channel input, denoted as $L_k$

\begin{figure*}[t]
  \centering
  \includegraphics[width=0.98\linewidth]{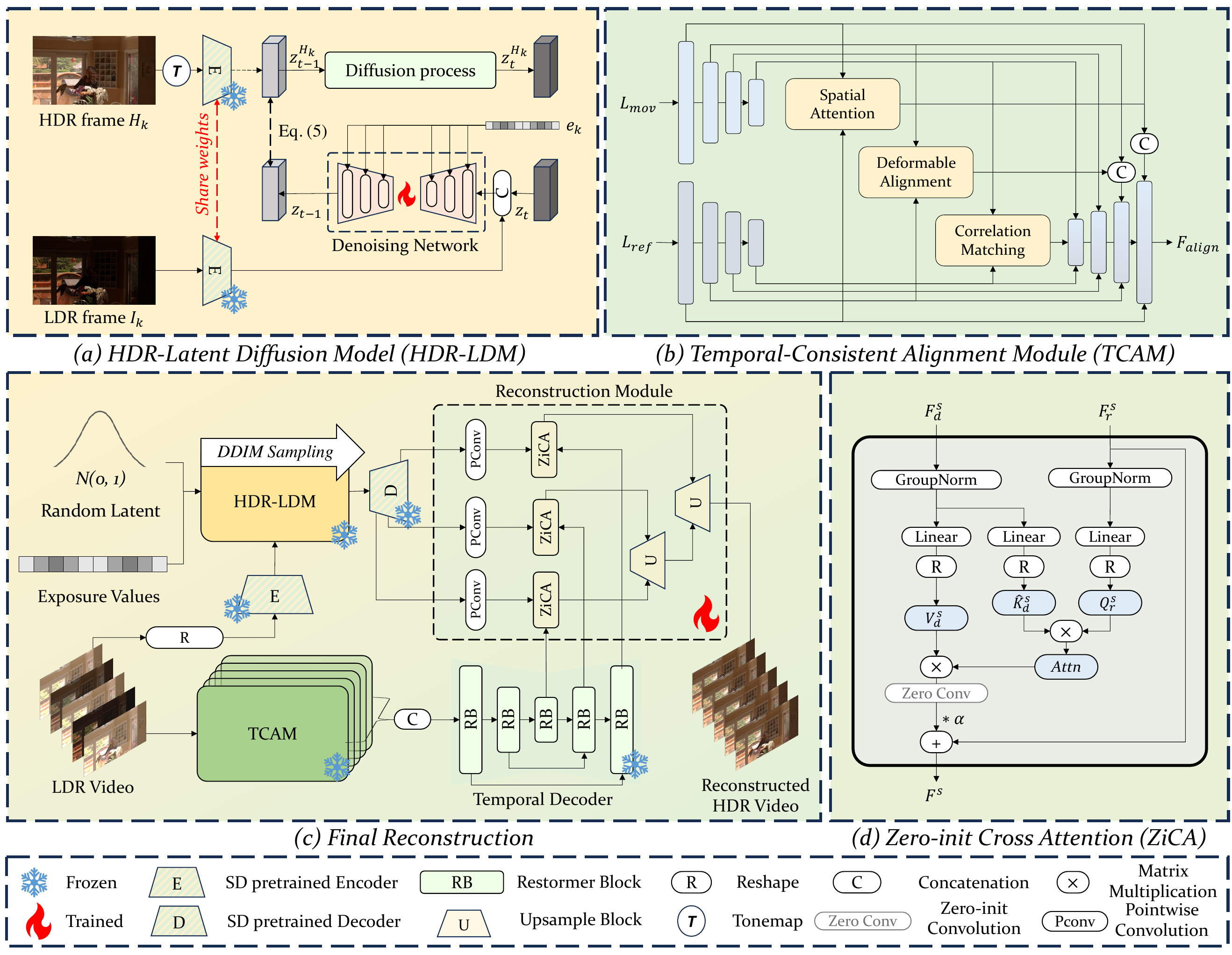}
  \vspace{-0.35cm}
   \caption{
   Pipeline of HDR-V-Diff comprises three key components:
   (a) depicts details about HDR-LDM, which aims to learn the distribution prior of single HDR frames. We introduce a tonemapping strategy to compress HDR frames into the latent space and a novel exposure embedding to integrate exposure information into the diffusion process.
   (b) depicts details about TCAM, which learns the temporal information by conducting coarse-to-fine feature alignment in different scales among video frames.
   (c) depicts details about the final reconstruction process, which leverages the proposed zero-init cross-attention mechanism to integrate the learned distribution prior and temporal information, achieving the high-quality HDR frames reconstruction.
   (d) illustrates the details of the proposed zero-init cross-attention mechanism.
   }
   \label{fig:main}
   \vspace{-0.22cm}
\end{figure*}

\vspace{-0.2cm}
\begin{equation}
L_k = c(I_k, I_k^\gamma/e_k),
\end{equation}

where $e_k$ is the exposure value of $I_k$, and $c(\cdot)$ denotes concatenation along the channel dimension. 
In this section, we present our HDR-V-Diff. As shown in Fig.~\ref{fig:main}, the proposed HDR-V-Diff consists of three main components: an HDR Latent Diffusion Model (HDR-LDM) for capturing the HDR distribution, presented in Sec.~\ref{sec:HDRLDM}; a Temporal-Consistent Alignment Module (TCAM) with a temporal decoder for inter-frame temporal information aggregation, presented in Sec.~\ref{sec:TCAM}; and a reconstruction module with a novel Zero-init Cross-Attention (ZiCA) to integrate the distribution prior and temporal information, depicted in Sec.~\ref{sec:final}. Due to the difference in the optimization target of each component, we respectively optimize the proposed components as depicted in Sec.~\ref{sec:final}.

\subsection{HDR Latent Diffusion Model}
\label{sec:HDRLDM}
Although regression-based methods can reconstruct most content in the HDR video frames and remove some degradations, the reconstructed frames often have obvious ghosting artifacts, random noise, and missing details in saturated regions, which are still far from the distribution of high-quality HDR video frames, as depicted in Fig.~\ref{fig:Teaser} (a) and (b).
HDR distribution prior would provide reliable guidance for the regression-based HDR video reconstruction methods, achieving a better trade-off between perception and fidelity. However, directly using video diffusion models would cause an unnecessary and massive computation burden since most pixels and information in LDR videos are known and repeated in every frame.
Therefore, we choose to leverage a single image LDM to estimate the HDR distribution prior and reduce demands in computation.

\textbf{Preliminary: Latent Diffusion} The proposed model is based on the paradigm of the LDM~\citep{rombach2022high} for efficiency, where the forward and reverse processes conducted on the latent space regulated by a pretrained auto-encoder~\citep{rombach2022high,razavi2019generating,yan2021videogpt}.
In the training phase, LDM methods first convert an image $x$ into a latent vector $z$ with the encoder $\mathrm{Enc}(\cdot)$, as $z = \mathrm{Enc}(x)$.
Then, in the forward process, gaussian noise at timestep $t$ with a variance of $\beta_{t} \in (0,1)$ is added to the encoded latent vector $z$ for producing the noisy latent. Each iteration of the forward process in latent space can be described as follows
\begin{equation}
    z_{t} = \sqrt{\alpha_{t}}z_{t-1} + \sqrt{1-{\alpha}_{t}}\epsilon_t,\label{eq:add_noise}
\end{equation}
where $\epsilon_t \sim \mathcal{N}(0,\textbf{I})$, $\alpha_{t} = 1- \beta_{t}$ and $z_{t}$ denotes noisy latent vector at $t$-th timestep. When $t$ increases, the latent $z_{t}$ gradually approximates a standard Gaussian distribution. In the backward process, the denoising network learns by predicting the added latent noise $\epsilon_t$ at a randomly selected timestep $t$, conditioning on the given condition $c^*$ (\textit{i.e.}, text prompts, semantic maps). The optimization of the denoising network can be formulated as follows
\begin{equation}
    \begin{aligned}
     \mathcal{L}_{ldm} = \mathbb E_{z,c^*,t,\epsilon}[||\epsilon_t -\epsilon_{\theta}(z_{t}, c^*, t)||^2_2], \label{eq:diff_loss}
    \end{aligned}
\end{equation}

where $x,c^*$ are sampled from the dataset, $t$ is uniformly sampled, and $\epsilon$ is sampled from the standard gaussian distribution. $\epsilon_{\theta}$ denotes the estimated noise from the denoising network.

\paragraph{Exposure Embedding}

\label{sec:exposureembedding}
In this specific task, the input LDR video frames are collected in an altering exposure way to capture more scene information in highlight and shadow regions.
The exposure time significantly affects the content of each frame as global information. Inspired by recent works~\citep{rombach2022high,mildenhall2021nerf}, we employ sinusoidal embeddings to encode the exposure value of the condition image as global information and insert it into the diffusion process
The calculation of the proposed exposure embedding $\boldsymbol{E}_{(e_k)}$ is formulated as follows


\vspace{-1mm}
\begin{equation}
    \boldsymbol{E}_{(e_k,2n)} = \operatorname{sin}\left(\frac{e_k}{{10000}^{2n/d}}\right), \quad \boldsymbol{E}_{(e_k,2n+1)} = \operatorname{cos}\left(\frac{e_k}{{10000}^{2n+1/d}}\right),
\end{equation}
\vspace{-1mm}

where $e_k$ denotes the exposure value of the $k$-th conditioned LDR image, and $d$ denotes the embedding dimension. The proposed exposure embedding is calculated and involved in the diffusion process in the same way as time embedding~\citep{stablediffusion}.
With the proposed exposure embedding, the denoising network can identify the exposure value of the conditioned latent from the embedded information, which allows for adaptive processing and the generalization of the network towards different capture settings (\textit{e.g.}, 2-exposures and 3-exposures). Ablation experiments about the proposed exposure embedding can be found in Table~\ref{tab:ablation}

\paragraph{Tonemapping Strategy} \label{sec:stage03}
The pretrained autoencoder of Stable Diffusion~\citep{stablediffusion} provides a strong capability and learns a robust discrete latent space, but the direct utilization of pretrained VAE failed due to the mismatch between the pretrained SDR dataset and the target HDR dataset. To obtain robust HDR latent representation, the $k$-th HDR frame $H_k$ is first tonemapped and then fed into the pretrained encoder to obtain their compressed latent representation $z^{H_k}$. The corresponding $k$-th LDR frame $I_k$ is directly fed into the pretrained encoder to obtain its latent representation $\mathrm{Enc}(I_k)$ as a condition.

For the denoising network, we employ the time-conditioned U-Net as in~\citep{rombach2022high}, but with two main modifications: 1) We insert the proposed exposure embedding $\boldsymbol{E}_{(e_k)}$ into the denoising process to enable adaptive generation given the altered-exposed LDR condition. 2) All the vanilla self-attention blocks are replaced with the MaxViT blocks~\citep{tu2022maxvit} to obtain computational efficiency.
By reformulated Eq.~\ref{eq:diff_loss}, the loss function for optimization is denoted as
\vspace{+1mm}
\begin{equation}
    \begin{aligned}
        \mathcal{L}_{hdrldm} = \mathbb E_{z,I_k,t,\epsilon_t}[||\epsilon_t-\epsilon_{\theta}(z_t, \mathrm{Enc}(I_k), \boldsymbol{E}_{(e_k)}, t)||^2_2]. \label{eq:hdrldm_diff_loss}
    \end{aligned}
\end{equation}

In summary, the proposed exposure embedding facilitates the ability of the HDR-LDM to generalize across various exposure settings, while the tonemapping strategy effectively transforms HDR content into latent space regulated by the pretrained autoencoders~\citep{stablediffusion}, thereby reducing computational overhead. The well-trained HDR-LDM is adept at estimating the distribution prior of HDR frames based on the provided LDR conditions. During the final reconstruction phase, the pretrained HDR-LDM is frozen and sampled using the DDIM strategy~\citep{song2020denoising}, with a sampling step set to 10 to strike a balance between efficiency and generation quality.

\subsection{Temporal-Consistent Alignment Module}
\label{sec:TCAM}
We propose TCAM to serve as a complement for HDR-LDM by exploiting the temporal information in the neighbor frames.
We present the detailed architecture of TCAM in Fig.~\ref{fig:main} (b). It takes the middle reference frame $L_{\text{ref}}$ and one moving frame $L_{\text{mov}}$ as inputs and produces the aligned feature $F_{align}$. The proposed TCAM utilizes shared weights across the entire input LDR frames. To optimize TCAM, we also employ a multi-scale temporal decoder constructed by restormer blocks~\citep{Restormer} to merge the aligned features from several TCAMs, as shown in Fig.~\ref{fig:main} (c).
 
Specifically, TCAM follows a U-shaped structure to conduct coarse-to-fine feature alignment at different scales, which enhances the robustness to large motions and reduces computational costs.
At the coarse scale, TCAM calculates the spatial attention~\citep{AHDRNet,hdrt} to selectively gather complementary information from the moving frame.
At the middle scale, TCAM leverages the advantages of the deformable convolution~\citep{edvr,iccv21hdr} to address moderate misalignment.
At the finest scale, TCAM performs patch correlation matching~\citep{TTSR,lanhdr} to identify the relevant patches between input frames. Finally, the multi-scale aligned features are gradually fused and restored to the original resolution.
\subsection{Final Reconstruction}
\label{sec:final}
To fully exploit the distribution prior from HDR-LDM and the temporal information from TCAM, we introduce a reconstruction module with a novel Zero-init Cross-Attention (ZiCA) mechanism to produce high-quality HDR frames.
As depicted in Fig.~\ref{fig:main} (c), the reconstruction module takes the multi-scale features generated by the pretrained stable diffusion decoder~\citep{stablediffusion} and the frozen temporal decoder introduced in Sec.~\ref{sec:TCAM}. For each scale, a corresponding ZiCA is introduced to integrate both features of that scale. The proposed ZiCA not only integrates the learned information, but also ensures a stable training process with the zero-init convolution. The integrated features of different scales are gradually upsampled and fused to produce the final HDR video frames.

\paragraph{Zero-init Cross-Attention (ZiCA)}


The architecture of ZiCA is shown in Fig.~\ref{fig:main} (d). Each ZiCA is fed with a distribution prior feature of scale $s$, denoted as $F_{d}^s$, and a temporal feature of the same scale, denoted as $F_{r}^s$.
$F_{d}^s$ is first mapped to the same channel dimension as $F_{r}^s$. Then, $F_{d}^s$ and $F_{r}^s$ are fed to the group-normalization and linear layers to aggregate cross-channel context.
After that, $F_{r}^s$ is selected to yield the query matrix $Q_{r}^s \in \mathbb{R}^{HW\times C}$, $F_{d}^s$ is selected to yield the key matrix $K_{d}^s \in \mathbb{R}^{HW\times C}$ and value matrix $V_{d}^s \in \mathbb{R}^{HW\times C}$.
Based on the generated matrices, we perform the cross-attention mechanism and then fed to a zero-init convolution $\mathcal{F}(\cdot;{\theta}^s)$, parameterized by ${\theta}^s$.
The output feature is multiplied by a predefined control scale $\alpha$ and added to the temporal feature $F_{r}^s$.
The whole process is formulated as

\vspace{-0.5em}
\begin{equation}
    F^s = F^s_r + \alpha \mathcal{F}( \operatorname{CrossAttention}(Q^s_r, K^s_d, V^s_d);\theta^s),
\end{equation}
\vspace{-1em}

where $F^s$ denotes the final output feature of ZiCA at scale $s$ and control scale $\alpha$ is set to 1. 
Specifically, the weight and bias parameters ${\theta}^s$ of the zero-init convolution layer are initialized to zero and progressively grow.
ZiCA can effectively avoid the optimization collapse caused by the representation gap between HDR-LDM and TCAM.
Ablation on ZiCA is provided in Table~\ref{tab:ablation}.

\paragraph{Training Strategy}
The target of HDR-LDM is to learn the distribution prior of HDR video frames, while the target of TCAM with the temporal decoder is to aggregate the complementary temporal information in the neighbor frames. The gap in learning targets leads to the difference in optimization. To construct and optimize the proposed HDR-V-Diff, we first optimized HDR-LDM with Eq.~\ref{eq:hdrldm_diff_loss} to capture the HDR distribution.

Then, TCAM with the temporal decoder is optimized in an end-to-end manner.
The loss function $\mathcal{L}_{TCAM}$ includes three terms: pixel-wise reconstruction loss $\mathcal{L}_1$ between the predicted HDR frame and the tonemapped ground truth HDR frame to ensure reconstruction quality, temporal loss $\mathcal{L}_{temp}$~\citep{lanhdr} to enhance the temporal consistency, and perceptual loss $\mathcal{L}_{perception}$~\citep{perloss} calculated based on pretrained VGG-19 network~\citep{VGG} to balance visual quality. The total loss is represented as

\vspace{-2mm}
\begin{equation}
\mathcal{L}_{TCAM} = \lambda_1 \mathcal{L}_1 + \lambda_2 \mathcal{L}_{temp} + \lambda_3 \mathcal{L}_{perception},  \label{eq:TCAM_loss}
\end{equation}
\vspace{-2mm}

where $\lambda_1$ is set to 1, $\lambda_2$ and $\lambda_3$ are set to 0.1. 

Finally, we freeze the parameters of HDR-LDM and TCAM with the temporal decoder, and optimize the reconstruction module to integrate the learned information from the two modules by Eq.~\ref{eq:TCAM_loss}.

\section{Experiments}

\paragraph{Experimental Setup}
Following the previous works~\citep{iccv21hdr,lanhdr}, we construct our synthetic training data based on the Vimeo-90K septuplet dataset~\citep{vimeo90k} by converting the original data to LDR video frames with alternating exposures. The proposed HDR-V-Diff is evaluated on the DeepHDRVideo dataset~\citep{iccv21hdr} and two synthetic videos (POKER FULLSHOT and CAROUSEL FIREWORKS) from the Cinematic Video dataset~\citep{cinematic14}. The DeepHDRVideo dataset~\citep{iccv21hdr} encompasses both real-world dynamic scenes and static scenes enhanced with random global motion. Additionally, the HDRVideo dataset~\citep{kala13} is exclusively employed for qualitative assessment, given the absence of ground truth. For evaluation, our results are compared with previous HDR video reconstruction methods~\citep{kala13,kala19,iccv21hdr,lanhdr} and state-of-the-art multi-exposure fusion methods~\citep{AHDRNet} after adapting them to the video reconstruction task. All the visual results are tonemapped using Photomatix and are best viewed by zooming into the electronic version. 
\paragraph{Evaluation metrics}
For an objective assessment, we compute PSNR$_T$, SSIM$_T$, and HDR-VDP-2 between the predicted results and ground truth frames. PSNR$_T$ and SSIM$_T$ are calculated on tonemapped images using the $\mu$-law~\citep{iccv21hdr,lanhdr,AHDRNet}. When computing HDR-VDP-2, we set the angular resolution of the image in terms of the number of pixels per visual degree as 30. Additionally, to evaluate the improvement of perceptual quality of the generated HDR images, we compute LPIPS~\citep{lpips} and MUSIQ~\citep{musiq} between the tonemapped predicted results and tonemapped ground truth frames.

\begin{figure*}[t]
  \centering
    
  \includegraphics[width=\linewidth]{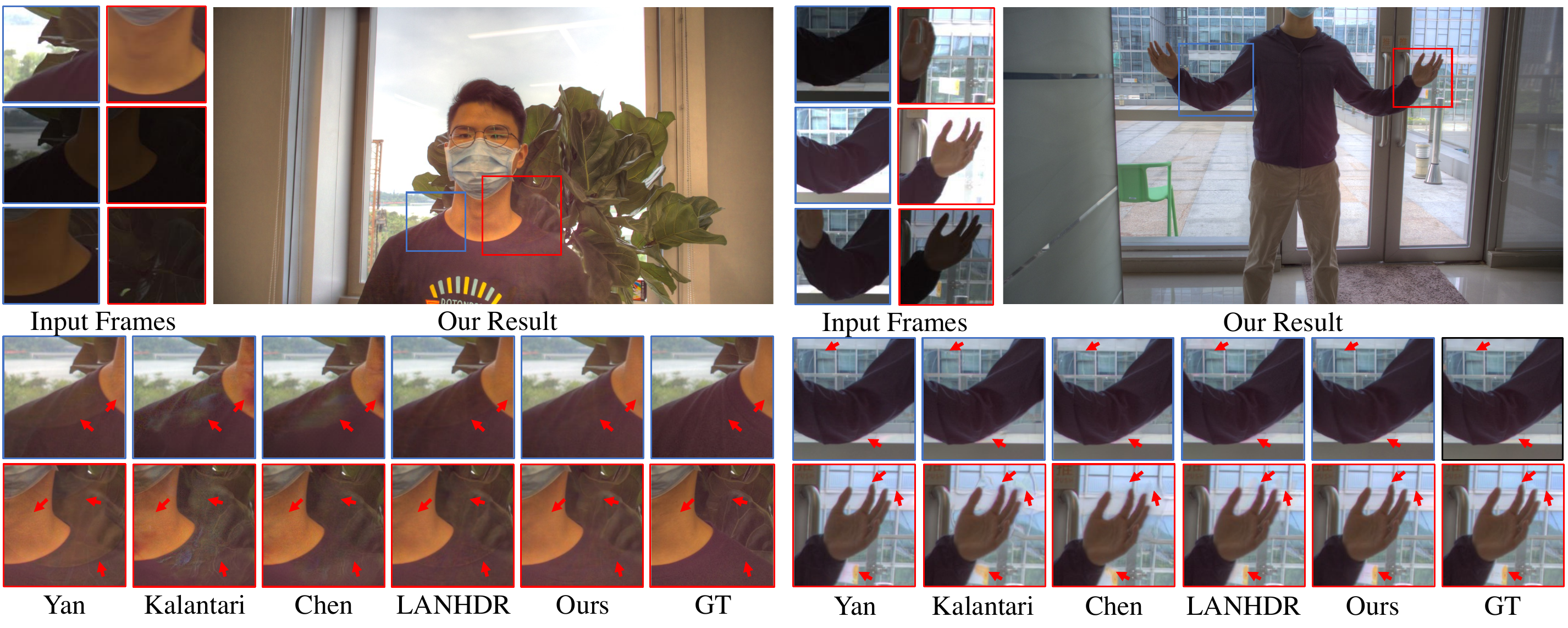}
   \vspace{-0.75cm}
   \caption{Visual comparisons of different methods on the video frames with 3-exposures from DeepHDRVideo dataset~\citep{iccv21hdr}.}
   \vspace{-0.15cm}
   \label{fig:deephdr}
\end{figure*}

\begin{table}[t] 
  \caption{Quantitative results of different methods averaged on the testing scenes from DeepHDRVideo dataset~\citep{iccv21hdr}. \textbf{Bold} indicates the best results and \underline{underline} indicates the second-best results.}
  \vspace{+0.2cm}
  \label{tab:deephdr_comapre_table}
  \centering
  \renewcommand{\arraystretch}{1} 
  \setlength{\tabcolsep}{3mm}
  \resizebox{0.98\linewidth}{!}{
  \begin{tabular}{l|cccccc}
    \toprule
    Methods (2-exposures) & $\text{PSNR}_T \uparrow$ & $\text{SSIM}_T \uparrow$ & LPIPS $\downarrow$ & MUSIQ $\uparrow$ & HDR-VDP-2 $\uparrow$ \\
    \midrule
    \citet{kala13} & 40.33 & 0.9409 & - & - & 66.11 \\
    \citet{kala19} & 39.91 & 0.9329 & 0.1238 & 50.92 & 71.11 \\
    \citet{AHDRNet} & 40.54 & 0.9452 & 0.1102 & 52.39 & 69.67 \\
    \citet{iccv21hdr} & \textbf{42.48} & \textbf{0.9620} & \underline{0.0971} & \underline{52.68} & \textbf{74.80} \\
    \citet{lanhdr} & 41.59 & 0.9472  & 0.1048 & 52.33 & \underline{71.34}\\
    Ours & \underline{42.07} & \underline{0.9604} & \textbf{0.0939}  & \textbf{52.71} & 70.88 \\
    \midrule
    Methods (3-exposures) & $\text{PSNR}_T \uparrow$ & $\text{SSIM}_T \uparrow$ & LPIPS $\downarrow$ & MUSIQ $\uparrow$ & HDR-VDP-2 $\uparrow$ \\
    \midrule
    \citet{kala13} & 38.45 & 0.9489 & - & - & 57.31 \\
    \citet{kala19} & 38.78 & 0.9331 & 0.1228 & 52.26 & 65.73 \\
    \citet{AHDRNet} & 40.20 & 0.9531 & 0.1062 & 53.84 & \underline{68.23}\\
    \citet{iccv21hdr} & 39.44 & \underline{0.9569} & 0.0917 & \underline{54.35} & 67.76 \\
    \citet{lanhdr} & \underline{40.48} & 0.9504 & \underline{0.0912} & 54.34 & \textbf{68.61}\\
    Ours & \textbf{40.82} & \textbf{0.9581} & \textbf{0.0822} & \textbf{54.55} & 68.16 \\
    \bottomrule
  \end{tabular}
  }
\end{table}

\begin{figure*}[!t]
  \centering
   \includegraphics[width=\linewidth]{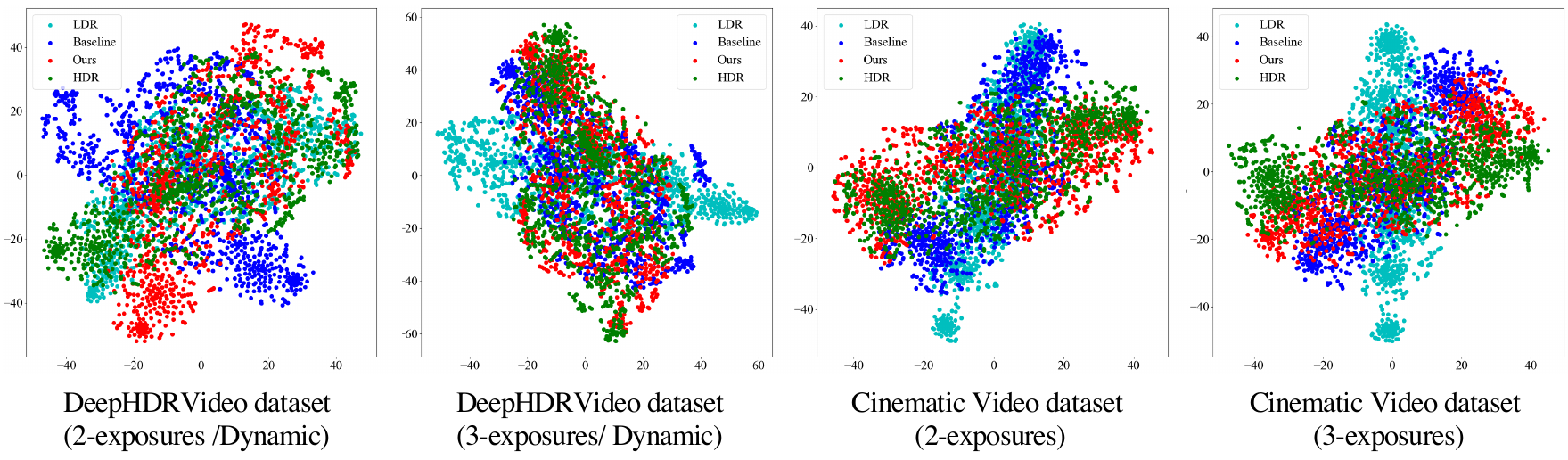}
   \vspace{-0.6cm}
   \caption{Analysis of distribution differences. We analyze the t-SNE distribution on the DeepHDRVideo dataset~\citep{iccv21hdr} and the Cinematic Video dataset~\citep{cinematic14}, including both 2-exposures and 3-exposures. We visualize the distribution of input LDR frames and HDR frames, the results of the regression-based baseline~\citep{lanhdr}, and our results. It can be observed that our results illustrate a closer distribution than the state-of-the-art regression-based method.}
   \vspace{-0.25cm}
   \label{fig:tsne_more}
\end{figure*}

\begin{figure*}[t]
  \centering
   \includegraphics[width=\linewidth]{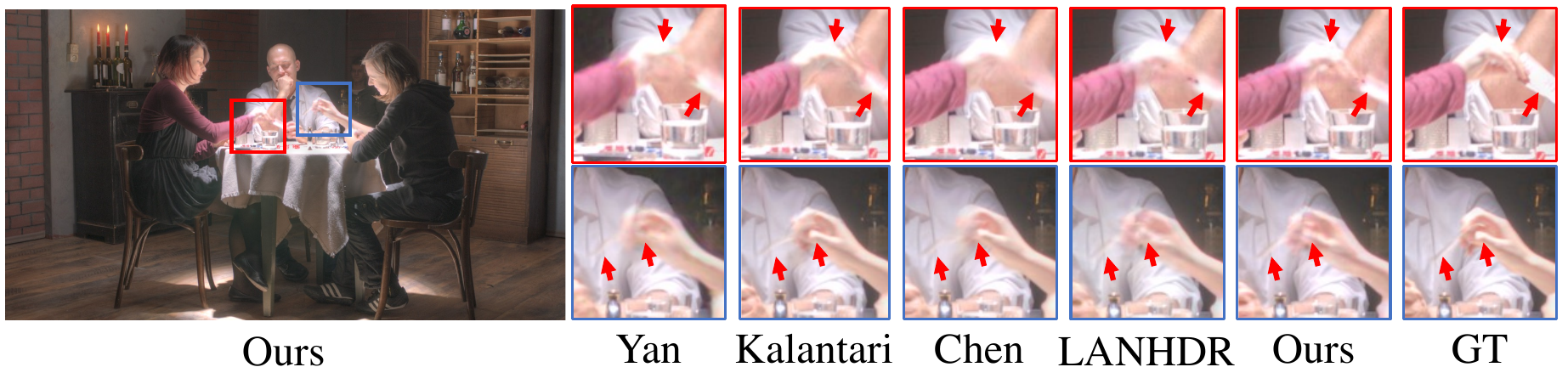}
   \vspace{-0.8cm}
   \caption{Visual comparisons of different methods on the video frames with 3-exposures from Cinematic Video dataset~\citep{cinematic14}.}
   \label{fig:cinematic}
   \vspace{-0.45cm}
\end{figure*}

\begin{table}[t]
  \caption{Quantitative results of different methods averaged on the testing scenes from Cinematic Video dataset~\citep{cinematic14}. \textbf{Bold} indicates the best results and \underline{underline} indicates the second-best results.}
  \vspace{+0.2cm}
  \label{tab:cinematic_comapre_table}
  \centering
  \renewcommand{\arraystretch}{1}  
  \setlength{\tabcolsep}{3mm}
  \resizebox{0.98\linewidth}{!}{
  \begin{tabular}{l|cccccc}
    \toprule
    Methods (2-exposures) & $\text{PSNR}_T \uparrow$ & $\text{SSIM}_T \uparrow$ & LPIPS $\downarrow$ & $\text{MUSIQ} \uparrow$ & $\text{HDR-VDP-2} \uparrow$ \\
    \midrule
    \citet{kala13} & 37.51 & 0.9016 & - & - & 60.16 \\
    \citet{kala19} & 37.06 & 0.9053 & 0.1201 & 40.53 & \underline{70.82} \\
    \citet{AHDRNet} & 31.65 & 0.8757 & 0.1364 & \textbf{42.43} & 69.05 \\
    \citet{iccv21hdr} & 35.65 & 0.8949 & \underline{0.1158} & 41.23 & \textbf{72.09} \\
    \citet{lanhdr} & \textbf{38.22} & \underline{0.9100} & 0.1163 & 41.14 & 69.15 \\
    Ours & \underline{37.86} & \textbf{0.9174} & \textbf{0.0986} & \underline{42.33} & 68.38 \\
    \midrule
    Methods (3-exposures) & $\text{PSNR}_T \uparrow$ & $\text{SSIM}_T \uparrow$ & LPIPS $\downarrow$ & $\text{MUSIQ} \uparrow$ & $\text{HDR-VDP-2} \uparrow$ \\
    \midrule
    \citet{kala13} & 30.36 & 0.8133 & - & - & 57.68\\
    \citet{kala19} & 33.21 & 0.8402 & 0.1496 & 41.43 & 62.44 \\
    \citet{AHDRNet} & 34.22 & 0.8604 & 0.1935 & 38.17 & \underline{66.18}\\
    \citet{iccv21hdr} & 34.15 & \underline{0.8847} & \underline{0.1418} & \underline{41.83} & \textbf{66.81}\\
    \citet{lanhdr} & \underline{35.07} & 0.8695 & 0.1814 & 38.54 & 65.42 \\
    Ours & \textbf{35.32} & \textbf{0.9051} & \textbf{0.1308} & \textbf{42.72} & 64.78\\
    \bottomrule
  \end{tabular}
  }
  \vspace{-0.55cm}
\end{table}

\begin{figure*}[t]
  \centering
   \includegraphics[width=\linewidth]{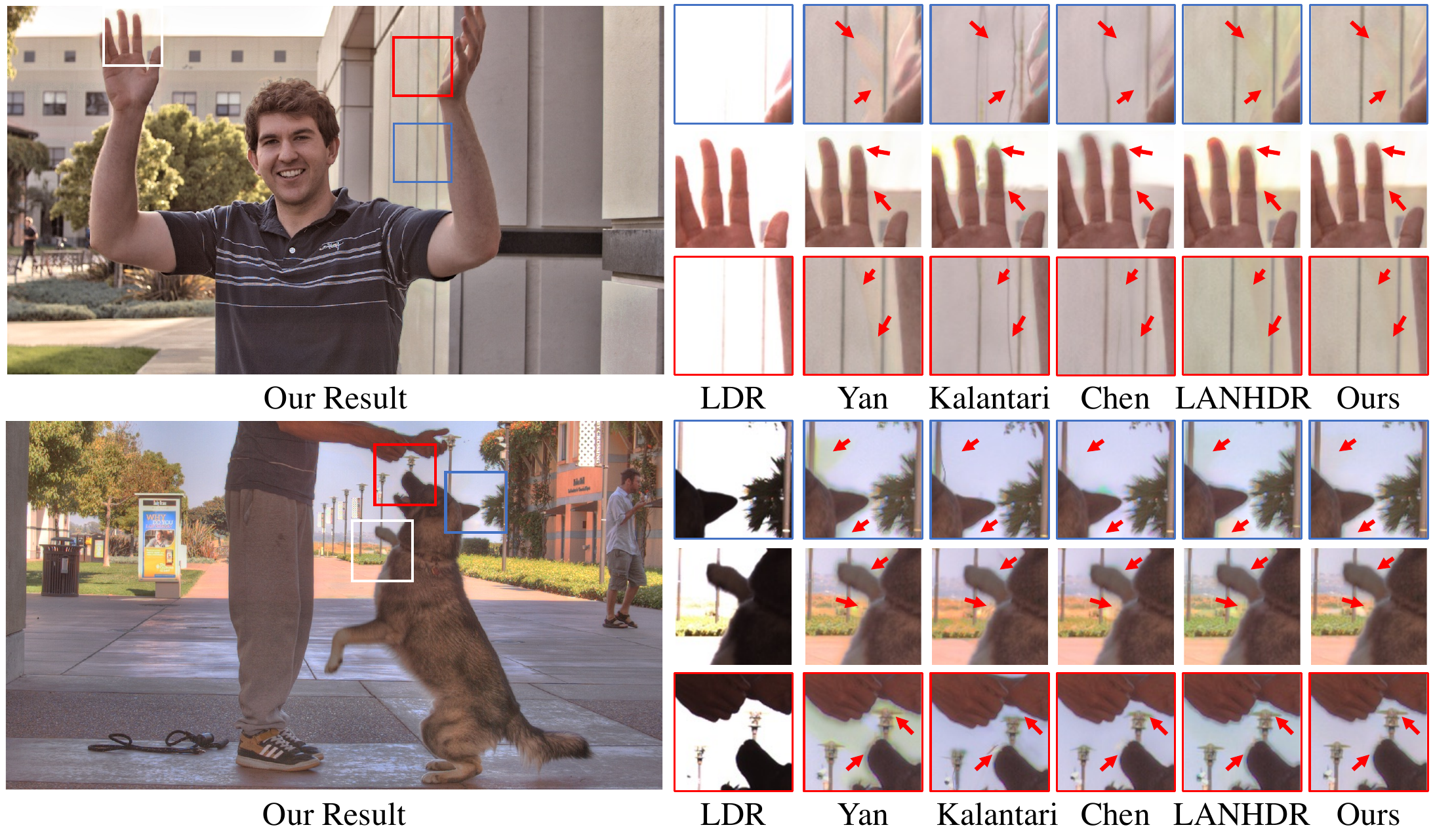}
   \vspace{-0.7cm}
   \caption{Visual comparisons of different methods on the video frames from HDRVideo dataset~\citep{kala13}.}
   \label{fig:hdrvideo}
   \vspace{-1.0cm}
\end{figure*}

\begin{table}[t]
\caption{Ablation on the proposed components.}
\label{tab:ablation}
  \centering
  \renewcommand{\arraystretch}{1}  
  \setlength{\tabcolsep}{3mm}{
  \resizebox{0.9\linewidth}{!}{
  \begin{tabular}{lcccccc}
    \toprule
Condition & Integration & PSNR$_\mu$ $\uparrow$ & SSIM$_\mu$ $\uparrow$ & LPIPS $\downarrow$ & MUSIQ $\uparrow$ \\
\midrule
w/o $\boldsymbol{E}_{(e_k)}$    & Concat   & 41.78   & 0.9566 & 0.1032 & 52.11 \\
w/o $\boldsymbol{E}_{(e_k)}$    & Cross-Attention & 41.86   & 0.9573 & 0.1021 & 52.41 \\
w/ $\boldsymbol{E}_{(e_k)}$ & Cross-Attention & {42.01} & {0.9583} & {0.0991} & {52.52} \\
w/ $\boldsymbol{E}_{(e_k)}$ & ZiCA    & {42.07}   & {0.9604} & {0.0939} & {52.71} \\
\bottomrule
  \end{tabular}}}
\end{table}
\vspace{-0.8cm}



\vspace{+1.3cm}
\paragraph{Quantitative Comparisons}
The comparison results of metrics on the DeepHDRVideo dataset~\citep{iccv21hdr} and Cinematic Video dataset~\citep{cinematic14} are respectively listed in Table \ref{tab:deephdr_comapre_table} and Table \ref{tab:cinematic_comapre_table}. It can be observed that our proposed HDR-V-Diff not only outperforms or matches previous state-of-the-art methods in objective metrics but also exhibits a significant improvement in subjective metrics. The improvement in subjective evaluation metrics correlates strongly with the improvement in the distribution of reconstructed results. Experimental results demonstrate that our method effectively reconstructs video frames that better fit in HDR distributions, as depicted in Fig.~\ref{fig:Teaser} (a) and Fig.~\ref{fig:tsne_more}. Each data point corresponds to a 256$\times$256 cropped patch. While the output of the baseline deviates from the distribution of HDR frames, our results exhibit a much closer alignment with ground truth. Findings from the regression-based baseline~\citep{lanhdr} reveal artifacts and noise due to its supervision strategy overlooking distribution learning, while the output of the proposed HDR-V-Diff demonstrates a more satisfactory distribution characteristic.

\paragraph{Qualitative Comparisons}
We visualize our results on video frames with significant exposure variance and noticeable large foreground motions. Fig.~\ref{fig:deephdr} illustrates the results on two 3-exposures video frames from the DeepHDRVideo dataset~\citep{iccv21hdr}. In those regions where fast motion occurs with exposure changes, previous methods fail to avoid ghosting artifacts and color distortions.
Flow-based methods~\citep{kala19,iccv21hdr} suffer from artifacts caused by flow estimation errors around the moving objects.
While the implicit feature alignment methods~\citep{AHDRNet,lanhdr} also contain unsuppressed artifacts from the misaligned neighbor frames.
In contrast, with the introduction and injection of the proposed diffusion prior, our results contain no obvious reconstruction error and maintain a satisfactory perceptual quality.
We also evaluate HDR-V-Diff on video frames with 3-exposures from the HDRVideo dataset~\citep{kala13}. Fig.~\ref{fig:hdrvideo} shows that our method produces the most reliable and satisfactory results, even when large occlusions and saturation exist.

\paragraph{Ablation Study}
We investigate the effectiveness of the proposed components, \textit{i.e.}, the exposure embedding in HDR-LDM and ZiCA, using 2-exposures videos from DeepHDRVideo dataset~\citep{iccv21hdr}. The ablation results are listed in Table~\ref{tab:ablation}. Specifically, condition indicates the condition mechanism of HDR-LDM, while integration indicates the combination method utilized in the reconstruction module. 
To evaluate the exposure embedding, we remove it and only input the LDR latent in the diffusion process.
To assess the proposed ZiCA module, we replace ZiCA with feature concatenation along the channel dimension and unmodified cross-attention~\citep{Restormer}.

As shown in Table~\ref{tab:ablation}, directly concatenates the distribution prior and temporal information affects both fidelity and perception quality compared to the second line.
When the injection mechanism becomes more adaptive with cross-attention, reconstruction quality notably improves. With the proposed exposure embedding, further quality enhancement is achieved, indicating that exposure embedding enhances the quality by conveying more global information about the latent LDR condition. Finally, the introduction of zero-init convolution leads to reasonable improvement by preserving the temporal information during the initial training. Gradually introducing the diffusion prior helps the entire network strike a better balance between fidelity and perception quality.

\section{Conclusion}
In this paper, we propose HDR-V-Diff, a new diffusion-promoted method for HDR video reconstruction, that leverages the diffusion model to capture the HDR distribution for generating high-quality results with realistic details.
To this end, we propose HDR-LDM to learn the distribution prior of single HDR frames and TCAM to learn temporal information as a complement for HDR-LDM, then design ZiCA to effectively integrate the learned information.
Extensive experiments validate that HDR-V-Diff achieves state-of-the-art results on several benchmarks.
Experiments also show the effectiveness and potential of diffusion models in video reconstruction tasks.

\bibliographystyle{iclr2024_conference}
\bibliography{iclr2024_conference}

\end{document}